\documentclass[10pt,twocolumn,letterpaper]{article}

\usepackage{cvpr}              

\usepackage{amsmath}
\usepackage{booktabs}
\usepackage{adjustbox}
\usepackage[table]{xcolor}
\usepackage{makecell}
\usepackage{arydshln}
\usepackage{colortbl}

\usepackage[accsupp]{axessibility} 

\definecolor{cvprblue}{rgb}{0.21,0.49,0.74}
\usepackage[pagebackref,breaklinks,colorlinks,allcolors=cvprblue]{hyperref}

\def\paperID{105} 
\def\confName{CVPR}
\def\confYear{2026}

\title{Graph-to-Frame RAG: Visual-Space Knowledge Fusion \\ for Training-Free and Auditable Video Reasoning}

\author{Songyuan Yang$^{1*}$ \quad Weijiang Yu$^{2\dagger}$ \quad Ziyu Liu$^2$ \quad Guijian Tang$^1$ \quad Wenjing Yang$^{1\dagger}$ \\ Huibin Tan$^{1*}$ \quad Nong Xiao$^2$ \\
$^1$National University of Defense Technology \qquad $^2$ Sun Yat-sen University
}

\begin{document}
\maketitle
\begin{abstract}
When video reasoning requires external knowledge, many systems with large multimodal models (LMMs) adopt retrieval augmentation to supply the missing context. Appending textual or multi-clip evidence, however, forces heterogeneous signals into a single attention space. We observe diluted attention and higher cognitive load even on non-long videos. The bottleneck is not only what to retrieve but how to represent and fuse external knowledge with the video backbone.
We present \textbf{Graph-to-Frame RAG (G2F-RAG)}, a training free and auditable paradigm that delivers knowledge in the visual space. On the offline stage, an agent builds a problem-agnostic video knowledge graph that integrates entities, events, spatial relations, and linked world knowledge. On the online stage, a hierarchical multi-agent controller decides whether external knowledge is needed, retrieves a minimal sufficient subgraph, and renders it as a single reasoning frame appended to the video. LMMs then perform joint reasoning in a unified visual domain. This design reduces cognitive load and leaves an explicit, inspectable evidence trail.
G2F-RAG is plug-and-play across backbones and scales. It yields consistent gains on diverse public benchmarks, with larger improvements in knowledge-intensive settings. Ablations further confirm that knowledge representation and delivery matter. G2F-RAG reframes retrieval as visual space knowledge fusion for robust and interpretable video reasoning.

\makeatletter{\renewcommand*{\@makefnmark}{}
\footnotetext{$^*$Equal contribution.
\textsuperscript{$^\dagger$}Corresponding author.}
}

\end{abstract}    
\begin{figure}[t]
    \centering
    \includegraphics[width=\linewidth]{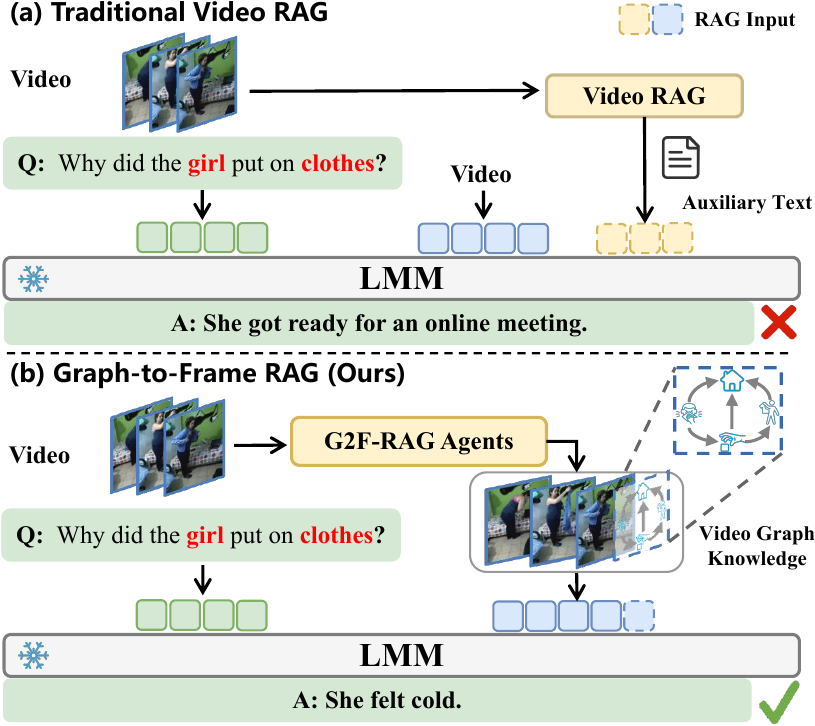}
    \vspace{-6.5mm}
    \caption{Traditional Video RAG appends auxiliary text, mixing heterogeneous tokens in one attention space, While G2F-RAG renders retrieved knowledge as a single reasoning frame appended to the video, keeping evidence in the visual space and producing more grounded causal answers.}
    \vspace{-7.5mm}
    \label{fig:introduction}
\end{figure}

\vspace{-4mm}
\section{Introduction}
\label{sec:intro}

Recent progress in large multimodal models (LMMs) \cite{qwen3vl, wang2025internvl3_5, gpt4o, lan2023learning} has advanced cross-modal representation, long-range temporal modeling, and open-domain reasoning, pushing video understanding from perception toward reasoning. Despite these gains, complex video reasoning still faces three fundamental challenges. First, compositional and multi-step reasoning \cite{yang2025vsibench}, such as cross-shot causality, navigation, and assembly, requires repeated alignment between local evidence and global structure. Second, external knowledge \cite{yang2025wildvideo} is often required, including commonsense, object functionality, and spatial topology, yet such information is not directly observable at the pixel level. Third, under constraints of computation and interpretability \cite{feng2025videor1}, we aim to solve complex problems reliably with small models without additional training, while providing an auditable evidence trail. This landscape motivates retrieval-augmented video understanding: retrieve information relevant to the question, then reason jointly with the video model. However, a key bottleneck remains how to feed external knowledge into the model with an appropriate way.

Mainstream video retrieval-augmented methods typically follow a retrieve-then-append paradigm with three common routes. The first \cite{luo2025videorag} appends textual context such as ASR, OCR, or descriptions to the prompt. The second \cite{ataallah2024goldfish, shen2025vgent} retrieves candidate clips or keyframes and provides them as additional visual inputs. The third \cite{ren2025videorag, hussein2019videograph} structures video into graphs or event chains, then injects the selected results as text or multi-segment context into the generator. These methods share an implicit assumption: obtaining more relevant content and providing a longer context will improve reasoning. In practice, we repeatedly observe performance degradation even when videos are short. Heterogeneous information sources must share a single attention space \cite{deng2025words}. Continuous low-level visual signals and discrete high-level text or dense visual text blocks compete for attention. The model must frequently switch and reconcile between two distinct representations. This leads to diluted attention and increased cognitive load \cite{shen2025vgent, liu2024lost}. The observation suggests that what to retrieve is important, but how to represent and fuse it with the backbone is equally crucial. When semantics are misaligned and the load is not controlled, retrieval can hinder the capability of models.

Guided by these observations, we shift the focus from extending videos or enlarging contexts to redesigning how knowledge is presented to the model. Video models are strongest when aggregating and reasoning within the visual space. External knowledge should therefore enter the same space with a visual grammar, which avoids cross-modal competition and reduces redundant tokens. Recent studies \cite{wei2025deepseek} also suggest that the visual modality can serve as an efficient compression medium for textual information, representing rich content with far fewer tokens. Inspired by this idea, we apply it to retrieval augmentation. Instead of appending text or many retrieved segments, we convert the retrieved structured knowledge into consumable visual tokens so that the model operates in its most familiar and well suited domain of spatiotemporal visual reasoning rather than in long context assimilation.

Building on this principle, we present Graph-to-Frame Retrieval-Augmented Generation (G2F-RAG) as a novel paradigm for delivering retrieved knowledge in the visual space, as shown in Figure~\ref{fig:introduction}. To instantiate this paradigm for video reasoning, we design a training-free two-stage multi-agent workflow, as shown in Figure~\ref{fig:method}. Offline, a graph-construction agent builds and caches a problem-agnostic, full video knowledge graph for each video. The graph covers entities, events, spatial relations, and essential links to external knowledge. It is constructed once and reused across questions. Online, an orchestration agent performs dynamic decisions via difficulty-aware hierarchical routing. If a question can be answered directly from the video evidence, a small LMMs answers it to preserve base performance. If a question requires external knowledge or multi-step reasoning, a retrieval agent retrieves a minimal sufficient reasoning chain or subgraph from the offline graph, a rendering agent renders it into a small number of lightweight and layout-controlled reasoning frames, and then feeds these frames together with the original video into the LMMs for joint reasoning. The reasoning frames are inserted or concatenated as real video frames so that temporal attention properly attends to them. The entire pipeline keeps the backbone frozen and changes only how knowledge is represented and fused, which addresses cross-modal competition and cognitive load at the source. Extensive experiments show that, under the same backbone, framing structured knowledge and fusing it in the visual modality significantly outperforms baselines. 

Our contributions are as follows: (1) To the best of our knowledge, we are the first to convert retrieved knowledge into the visual modality for video reasoning. Specifically, we propose Graph-to-Frame Retrieval-Augmented Generation (G2F-RAG), which renders retrieved knowledge into a video frame and feeds it as part of the video, mitigating cognitive load and retrieval-context explosion at the source. (2) To instantiate this paradigm, we design a training-free, two-stage multi-agent workflow: offline a graph-construction agent builds and caches a reusable, problem-agnostic video knowledge graph; online a controller performs difficulty-aware routing, retrieves a minimal subgraph, and renders a single layout-controlled reasoning frame appended to the video for joint inference with frozen LMM backbones. (3) We demonstrate strong performance and generalization across benchmarks and backbones. Comprehensive analyses reveal a central conclusion: compared to unbounded context expansion, delivering necessary knowledge as visual frames is better aligned with the inductive biases of video models for retrieval-augmented reasoning.

\begin{figure*}[t]
    \centering
    \includegraphics[width=\linewidth]{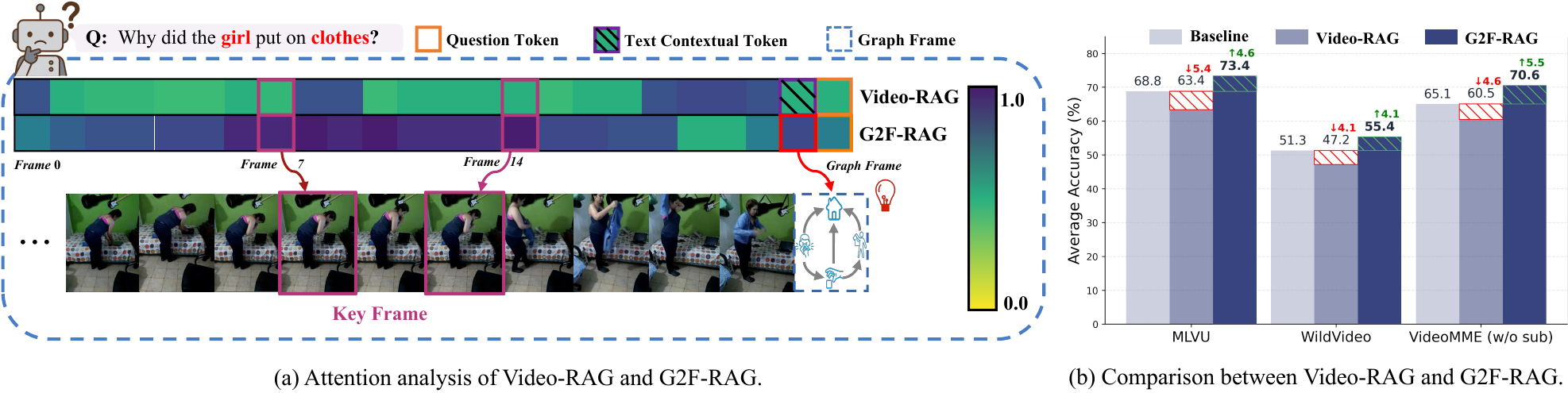}
    \vspace{-6mm}
    \caption{(a) Attention Analysis. Text-append Video-RAG diverts attention from key frames toward contextual tokens, while G2F-RAG keeps focus on key frames and the appended graph frame. (b) Comparison of the performance. Video-RAG causes consistent drops, whereas G2F-RAG yields clear gains over the baseline on all three benchmarks.}
    \label{fig:attention}
    \vspace{-5mm}
\end{figure*}

\vspace{-1mm}
\section{Related Work}
\vspace{-1mm}
\label{sec:formatting}

\subsection{Large Multimodal Models}
Building on the success of LLMs \cite{vicuna, yi, qwen3, llama4, gpt4}, Large Multimodal Models (LMMs) have advanced visual language understanding rapidly~\cite{jiang-etal-2025-specific}. LLaVA \cite{liu2023llava, liu2023improvedllava, liu2024llavanext} popularized instruction tuning on GPT-4 \cite{gpt4} curated data and established a widely adopted recipe for constructing LMMs, inspiring many follow up systems \cite{zhou2024tinyllava, caffagni2024wiki, cai2025llava}. Fueled by short-video applications and video generation, video understanding \cite{zhang2023videollama, yao2024minicpm, zhang2024llavavideo} has emerged as a focal area, yet spatio-temporal joint modeling remains challenging, demanding fine-grained spatial grounding and multi-hop reasoning, especially in long-video settings \cite{chen2024longvila, li2024videochatflash, zhang2025deep, pang2025mr}. To improve efficiency and scalability, Vamba \cite{ren2025vamba} introduces a hybrid Mamba–Transformer architecture with near-linear token complexity, while recent efforts \cite{feng2025videor1, li2025star, yan2025videochatr15} explore GRPO-based reinforcement fine-tuning for video LMMs and agentic pipelines for planning and tool use. In practice, closed-source commercial systems (e.g., GPT-4o \cite{gpt4o}, Gemini \cite{gemini1.5}) currently lead on video understanding, whereas open-source families such as Qwen-VL Series \cite{Qwen-VL, Qwen2-VL, Qwen2.5-VL, qwen3vl}, InternVL Series \cite{chen2024internvl, chen2024internvl2_5, zhu2025internvl3, wang2025internvl3_5}, and VideoLLaMA Series \cite{zhang2023videollama, cheng2024videollama2, zhang2025videollama3} provide widely used, reproducible research baselines. 
However, current LMMs still lack a principled way to ingest external knowledge without diluting attention or requiring retraining, and our Graph-to-Frame RAG addresses this by fusing knowledge in the visual space for training-free, auditable reasoning.

\begin{figure*}[t]
    \centering
    \includegraphics[width=\linewidth]{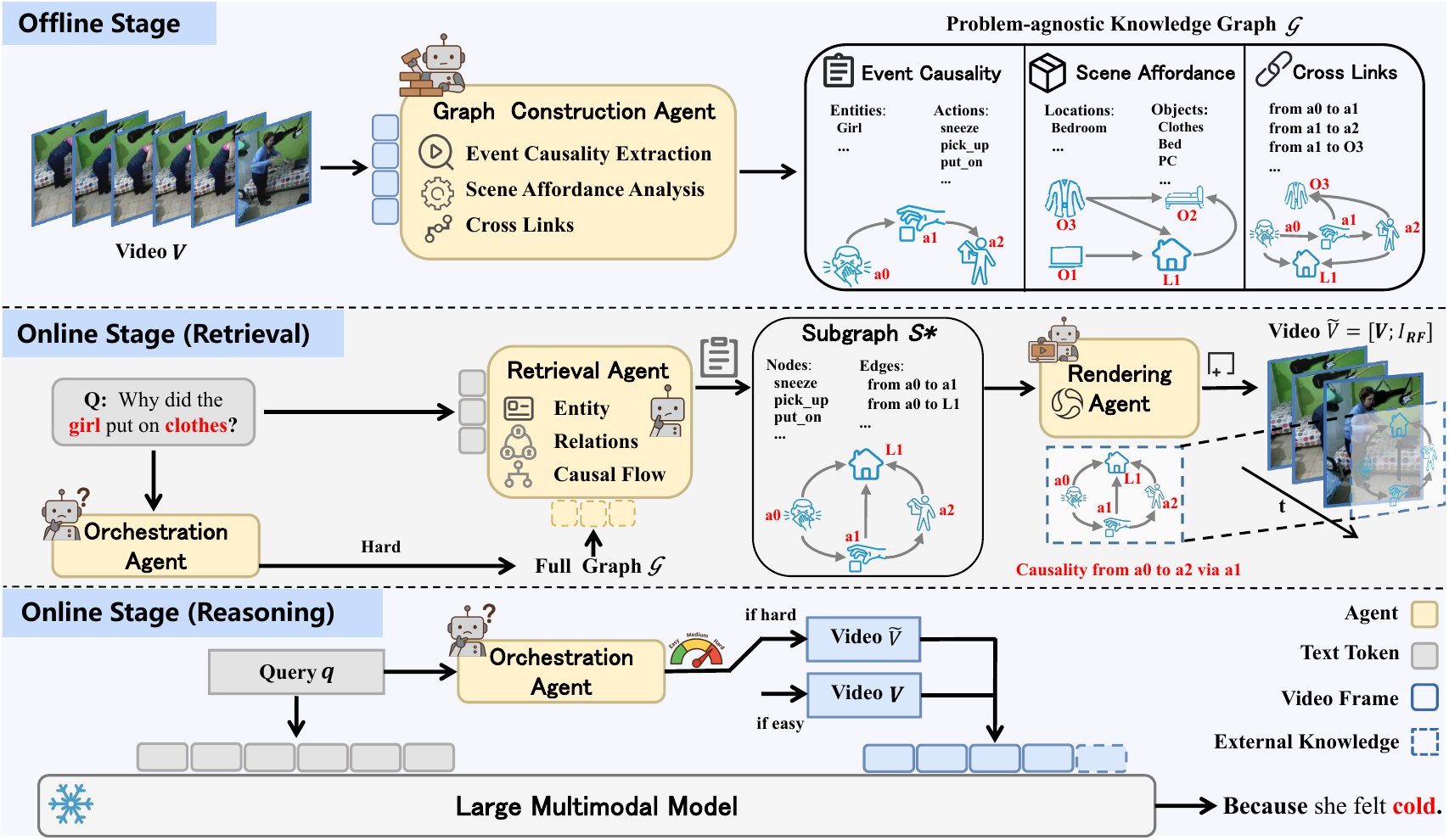}
    \vspace{-6mm}
    \caption{\textbf{Method overview.} Offline: a graph-construction agent builds a problem-agnostic video knowledge graph with optional external knowledge. Online: an orchestration agent routes the query; a retrieval agent extracts a minimal subgraph $S^*$; a rendering agent converts it into one reasoning frame and appends it to the video to form $\tilde{V}=[V; I_{\mathrm{RF}}]$. Finally, the LMM answers from $V$ for easy cases, or from $\tilde{V}$ for hard cases. The pipeline is training free, auditable, and fuses knowledge in the visual space.}
    \label{fig:method}
    \vspace{-5mm}
\end{figure*}

\subsection{Video Augmented Understanding}
Retrieval-Augmented Generation (RAG) is a primary paradigm for Video Augmented Understanding \cite{ren2025videorag, jeong2025videorag, luo2025videorag} in LMMs. It \cite{li2024structrag, gao2023retrieval} indexes clips, subtitles, and structured cues, retrieves evidence at query time, and conditions generation on the retrieved context, which improves factual grounding, reduces hallucination, and mitigates long context limits. Progress \cite{ataallah2024goldfish, han2024retrieval, jeong2025videorag} spans finer retrieval units, learned aggregation across clips, and knowledge representations that range from chunk based stores to graph structures, sometimes with adaptive routing that balances accuracy and efficiency. Remaining challenges \cite{,ren2025videorag, shen2025vgent, xue2025adavideorag} include temporal dispersion in long video, multi hop dependencies, retrieval noise, and the cost of constructing and maintaining large structured memories. Complementing retrieval, recent Video Augmented Understanding introduces light visual scaffolds that make temporal and structural information explicit, for example numbering frames \cite{wu2025number}, drawing progress indicators to surface relevant segments \cite{zhang2025vtimecot}, or translating images into symbolic SVG for compositional reasoning \cite{lin2025vcodemultimodalcodingbenchmark}. 
However, appending retrieved text or multi-clip evidence often creates attention competition and amplifies noise, and our approach resolves this by converting the retrieved subgraph into a single reasoning frame for same-modality fusion.

\section{Method}

\subsection{Attention Analysis}
Current video LMMs encode a video as frame tokens and align them with query tokens in a single attention space. Traditional text-based video RAG methods appends retrieved passages as additional tokens. Using Qwen-2.5VL \cite{Qwen2.5-VL} as baseline, we aggregate the attention scores from all the visual tokens corresponding to each frame across all attention heads and separate query tokens from retrieved-context tokens. In Figure~\ref{fig:attention}.(a), text-based Video-RAG \cite{luo2025videorag} disperses attention toward the long retrieved context and non-salient frames, reducing focus on key frames. With our G2F-RAG, attention concentrates on the key segment and on the single graph frame appended at the end.

The quantitative trends in Figure~\ref{fig:attention}.(b) mirror this behavior. On MLVU  \cite{zhou2025mlvu}, Video-RAG reduces accuracy by 5.4 points relative to the baseline, while G2F-RAG improves by 4.6 points over the baseline and by 10.0 points over Video-RAG. On WildVideo \cite{yang2025wildvideo}, the drops and gains are 4.1 and 3.9 points. On VideoMME \cite{fu2025video}, they are 4.6 and 5.5 points. Appending retrieved text consistently harms performance, whereas delivering the same knowledge as a single reasoning frame in the visual domain yields stable gains under the same backbone. The key factor is not only what to retrieve, but how to represent and fuse it with the video.

\subsection{Multi-Agent Framework}

We present a training free multi-agent framework that fuses external knowledge with video in a single visual domain. As shown in Figure~\ref{fig:method}, The workflow has an offline stage and an online stage. In the offline stage, a networked graph-construction agent generates and caches a full knowledge graph for each video. In the online stage, an orchestration agent receives a query, performs difficulty-aware routing, invokes a retrieval agent to extract a minimal sufficient subgraph, and invokes a rendering agent to convert this subgraph into a single reasoning frame. The video and the reasoning frame are then provided to LMMs for joint reasoning and answer generation.

Formally, given a video $V$ and a query $q$, the offline graph-construction agent is denoted by $\mathsf{Build}$ and outputs a problem-agnostic knowledge graph $\mathcal{G}=\mathsf{Build}(V)$. The graph $\mathcal{G}=(\mathcal{V},\mathcal{E})$ covers entities, events, actions, spatial locations, affordances, and concept-level knowledge, with cross-layer links. In the online stage, an orchestration agent $\mathsf{Route}$ decides a path $\mathsf{Route}(q,V,\mathcal{G})\in{\text{easy},\text{hard}}$. If the decision is easy, LMMs answer directly on the video to preserve base performance. If the decision is hard, a retrieval agent selects a compact subgraph $S^\star$ from $\mathcal{G}$ that maximizes relevance under a complexity budget:
\begin{equation}
S^\star=\arg\max_{S\subseteq \mathcal{G}} \left[ R(q,S) - \lambda,C(S) \right],
\end{equation}
where $R(q,S)$ measures relevance, $C(S)$ regularizes complexity, and $\lambda>0$ balances the two terms. A rendering agent converts $S^\star$ into a single high-readability reasoning frame $I_{\mathrm{RF}}=\mathsf{Render}(S^\star)$. We append $I_{\mathrm{RF}}$ as the last frame of the video to obtain the fused input $\tilde{V}=[V; I_{\mathrm{RF}}]$. LMMs then produce the final answer $a=\mathsf{LMMs}(\tilde{V}, q)$. The framework keeps the backbone frozen and changes only how knowledge is represented and delivered. It preserves performance on easy cases and unlocks the benefits of modality-aligned fusion for complex reasoning.

\begin{figure}[t]
    \centering
    \includegraphics[width=\linewidth]{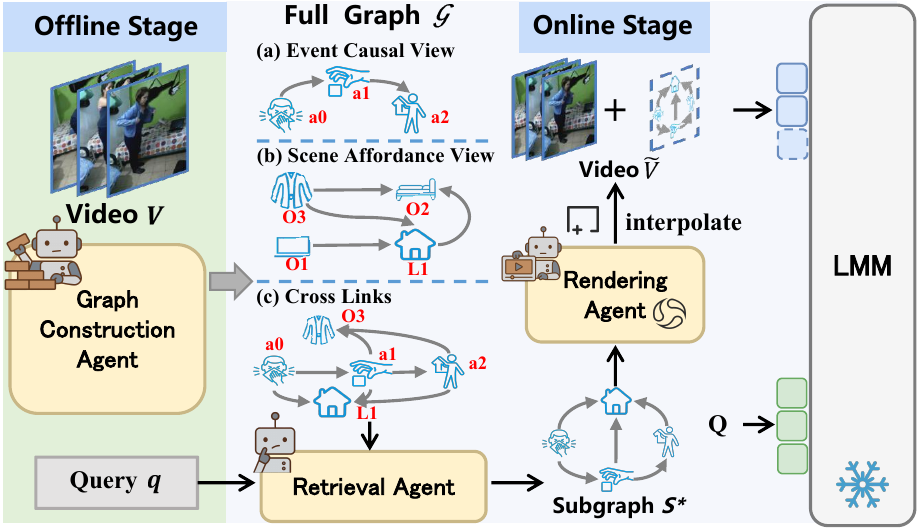}
    \vspace{-6mm}
    \caption{\textbf{G2F-RAG pipline.} Build a full video graph, retrieve a minimal subgraph for the query, render it as a single frame, and append it to the video for visual-space reasoning with the LMM.}
    \label{fig:graph2frame}
    \vspace{-7mm}
\end{figure}

\subsection{Graph-to-Frame Knowledge Fusion}

Graph-to-frame fusion is the core of the method, as shown in Figure \ref{fig:graph2frame}. In the offline stage, a networked graph-construction agent analyzes the video under a curated prompt and produces a queryable full graph $\mathcal{G}$. The design unifies two complementary views. The event–causal view models participants, actions, intent, preconditions, postconditions, and event-level abstractions connected by causal links. The scene–affordance view models objects and their affordances, functional areas and connectivity, and abstract concept knowledge. Dense cross-links bind the two views, which allows seamless transitions between “what happened” and “where or with what it happened.” The resulting graph covers visible facts and attaches essential world knowledge. It provides a reliable substrate for downstream retrieval and reasoning.

In the online stage, the retrieval agent maps the natural-language query $q$ to a structured subgraph query over $\mathcal{G}$ and returns a compact subgraph $S^\star$ that is minimal yet sufficient for the target reasoning. We impose explicit complexity constraints to control the visual token budget after rendering, such as $|\mathcal{V}(S^\star)|\le N_{\max}$ and $|\mathcal{E}(S^\star)|\le E_{\max}$. A rendering agent then converts $S^\star$ into a single reasoning frame $I_{\mathrm{RF}}$ that uses clear visual grammar to depict key entities, relations, and causal flow. The frame employs simple geometric primitives and short labels to ensure high readability and low noise within the visual domain. The reasoning frame does not encode temporal spans or timestamps. It focuses on structure and mechanism, which helps LMMs internalize external knowledge without being distracted by redundant details.

We append $I_{\mathrm{RF}}$ as the last frame to form $\tilde{V}=[V; I_{\mathrm{RF}}]$. Placing the reasoning frame at the end avoids interference with the model’s learning of the original content while ensuring that temporal attention can still attend to the inserted frame. A concise instruction enforces consistency, with original video context treated as authoritative. Conceptually, the process maps “retrieved structured knowledge” to “a small number of visual tokens” that LMMs can consume efficiently.

\subsection{Hierarchical Mechanism}
The hierarchical mechanism uses an agent to classify question difficulty and route computation accordingly. The goal is to protect performance on easy cases while directing complex cases to the knowledge-enhanced, modality-aligned path. Given $(q,V,\mathcal{G})$, the agent outputs a binary decision $d(q,V,\mathcal{G})\in{\text{easy},\text{hard}}$. We view the decision as an approximate utility comparison. The hard branch is chosen when the expected utility of graph-to-frame fusion exceeds that of direct video reasoning by a safe margin:
\begin{equation}
d(q,V,\mathcal{G})=
\begin{cases}
\text{hard}, & \Delta U(q,V,\mathcal{G}) \ge \tau,\\
\text{easy}, & \text{otherwise}.
\end{cases}
\end{equation}

\begin{equation}
\Delta U(q,V,\mathcal{G})=\widehat{U}{\mathrm{G2F}}(q,V,\mathcal{G})-\widehat{U}{\mathrm{Base}}(q,V).
\end{equation}
Here $\widehat{U}{\mathrm{G2F}}$ and $\widehat{U}{\mathrm{Base}}$ are proxy scores produced by the agent under different prompts. They summarize factors such as locality of evidence, cross-scene dependency, external concept triggers, and estimated reasoning steps. The threshold $\tau$ acts as a guard to protect easy-case performance. The agent also exposes uncertainty and a fallback signal. If confidence is low or the selected subgraph exceeds the budget, the system falls back to the easy branch or retries with tighter constraints.

The mechanism does not learn new parameters. It relies on prompt design to let the agent perform task decomposition and strategy selection in the input space. Since the decision depends on $(q,V,\mathcal{G})$ and on explicit detection of external knowledge needs, it favors problems that require structured, cross-entity, or cross-location reasoning and routes them to retrieval and graph-to-frame fusion. For questions that can be answered directly from the video, it keeps the shortest path and avoids unnecessary knowledge injection. Combined with the multi-agent orchestration and graph-to-frame fusion, the hierarchical mechanism achieves “direct for easy, enhanced for hard” without modifying LMMs.

\section{Experiments}

\begin{table*}[h]
\centering
\caption{\textbf{Performance comparison on sota LMMs.} All scores are average accuracy (\%) with 3 significant digits, except MMBench-Video which is graded by an LLM evaluator. G2F-RAG's absolute gains on each baseline are shown in superscripts.}
\vspace{-1.5mm}
\begin{adjustbox}{width=\linewidth,center}
\renewcommand{\arraystretch}{1.2}
\setlength{\tabcolsep}{1.5mm}
\begin{tabular}{lllllllll}
\toprule
& & \multicolumn{3}{c}{\textbf{General Multi Task}} & \multicolumn{2}{c}{\textbf{Temporal Spatial Reasoning}} & \multicolumn{2}{c}{\textbf{Open World}} \\
\cmidrule(lr){3-5}\cmidrule(lr){6-7}\cmidrule(l){8-9}
\textbf{Model} & \textbf{Size} & \textbf{MVBench} & \makecell{\textbf{VideoMME} \\ \textbf{(w/o sub)}} & \textbf{MMBench-Video} & \textbf{TempCompass} & \textbf{VSIBench} & \textbf{WildVideo} & \textbf{VideoMMMU} \\
\midrule
\multicolumn{9}{c}{\textbf{\textit{Proprietary LMMs}}} \\
\midrule
Gemini1.5 Pro \cite{gemini1.5} & - & 60.5 & 77.4 & 1.80 & 68.4 & 48.8 & 53.7 & 53.5 \\
GPT-4o\textsubscript{ (2024-05-13)} \cite{gpt4o}  & - & 57.5 & 67.9 & 1.87 & 69.5 & 34 & 62.1 & 61.2 \\
\midrule
\multicolumn{9}{c}{\textbf{\textit{Open-Source LMMs (5B$<$)}}} \\
\midrule
Qwen2.5-VL \cite{Qwen2.5-VL} & 3B & 67.0 & 61.5 & 1.63 & 64.4 & 27.9 & 38.6 & 42.3\\
Qwen3-VL \cite{qwen3vl} & 4B & 68.9 & 69.3 & - & - & 58.4 & - & 56.2 \\
\hdashline
InternVL3.5 \cite{wang2025internvl3_5} & 4B & 71.2 & 65.4 & 1.59 & 65.4 & 54.9 & 45.2 & 58.3 \\
\rowcolor{blue!7} ~~~~+ G2F-RAG (Ours) & 4B & 74.8\textcolor{blue}{\textsuperscript{+3.6}} & 70.1\textcolor{blue}{\textsuperscript{+4.7}} & 1.66\textcolor{blue}{\textsuperscript{+0.07}} & 68.7\textcolor{blue}{\textsuperscript{+3.3}} & 59.0\textcolor{blue}{\textsuperscript{+4.1}} & 47.1\textcolor{blue}{\textsuperscript{+1.9}} & 62.4\textcolor{blue}{\textsuperscript{+3.9}} \\
\midrule
\multicolumn{9}{c}{\textbf{\textit{Open-Source LMMs ($>$5B, 10B$<$)}}} \\
\midrule
InternVideo2.5 \cite{wang2025internvideo25} & 7B & 75.7 & 65.1 & - & - & - & - & 43.0 \\
VideoLLaMA 3 \cite{zhang2025videollama3} & 7B & 69.7 & 61.0 & - & - & - & - & 46.0 \\
VILA-1.5 \cite{lin2023vila} & 8B & - & - & - & 58.8 & 28.9 & - & 20.9 \\
MiniCPM-V 2.6 \cite{yao2024minicpm} & 8B & 44.7 & 59.7 & 1.70 & 59.6 & - & 46.4 & - \\
MiniCPM-V 4.5 \cite{yu2025minicpm45} & 8B & - & 67.9 & - & - & - & - & - \\
InternVL2.5 \cite{chen2024internvl2_5} & 8B & 70.5 & 63.7 & 1.68 & 68.7 & 41.6 & - & -\\
InternVL3 \cite{zhu2025internvl3} & 8B & 73.2 & 66.0 & 1.69 & 70.4 & 41.6 & - & 48.9\\
Qwen3-VL \cite{qwen3vl} & 8B & 68.7 & 71.4 & - & - & 59.4 & - & 65.3 \\
\hdashline
LLaVA-Video \cite{zhang2024llavavideo} & 7B & 62.1 & 63.7 & 1.60 & 65.5 & 35.7 & 53.4 & 36.1\\
\rowcolor{blue!7} ~~~~+ G2F-RAG (Ours) & 7B & 68.5\textcolor{blue}{\textsuperscript{+6.4}} & 64.5\textcolor{blue}{\textsuperscript{+0.8}} & 1.66\textcolor{blue}{\textsuperscript{+0.06}} & 69.0\textcolor{blue}{\textsuperscript{+3.5}} & 42.5\textcolor{blue}{\textsuperscript{+6.8}} & 57.0\textcolor{blue}{\textsuperscript{+3.6}} & 44.0\textcolor{blue}{\textsuperscript{+7.9}}\\
Qwen2.5-VL \cite{Qwen2.5-VL} & 7B & 67.5 & 65.1 & 1.79 & 69.2 & 37.4 & 51.3 & 47.4\\
\rowcolor{blue!7} ~~~~+ G2F-RAG (Ours) & 7B & 73.2\textcolor{blue}{\textsuperscript{+5.7}} & 70.6\textcolor{blue}{\textsuperscript{+5.5}} & 1.85\textcolor{blue}{\textsuperscript{+0.06}} & 72.0\textcolor{blue}{\textsuperscript{+2.8}} & 44.8\textcolor{blue}{\textsuperscript{+7.4}} & 55.4\textcolor{blue}{\textsuperscript{+4.1}} & 53.1\textcolor{blue}{\textsuperscript{+5.7}}\\
InternVL3.5 \cite{wang2025internvl3_5} & 8B & 72.1 & 66.0 & 1.74 & 71.0 & 56.3 & 53.0 & 63.4 \\
\rowcolor{blue!7} ~~~~+ G2F-RAG (Ours) & 8B & 78.5\textcolor{blue}{\textsuperscript{+6.4}} & 72.0\textcolor{blue}{\textsuperscript{+6.0}} & 1.82\textcolor{blue}{\textsuperscript{+0.08}} & 75.5\textcolor{blue}{\textsuperscript{+4.5}} & 60.1\textcolor{blue}{\textsuperscript{+3.8}} & 60.1\textcolor{blue}{\textsuperscript{+7.1}} & 68.2\textcolor{blue}{\textsuperscript{+4.8}}\\
\midrule
\multicolumn{9}{c}{\textbf{\textit{Open-Source LMMs ($>$10B)}}} \\
\midrule
InternVL3 \cite{zhu2025internvl3} & 14B & 76.6 & 70.4 & 1.73 & - & 48.9 & - & - \\
\hdashline
InternVL3.5 \cite{wang2025internvl3_5} & 14B & 72.8 & 67.9 & 1.73 & 71.2 & 60.8 & 60.1 & 65.0 \\
\rowcolor{blue!7} ~~~~+ G2F-RAG (Ours) & 14B & 77.4\textcolor{blue}{\textsuperscript{+4.6}} & 70.6\textcolor{blue}{\textsuperscript{+2.7}} & 1.80\textcolor{blue}{\textsuperscript{+0.07}} & 76.8\textcolor{blue}{\textsuperscript{+5.6}} & 62.9\textcolor{blue}{\textsuperscript{+2.1}} & 63.2\textcolor{blue}{\textsuperscript{+3.1}} & 71.2\textcolor{blue}{\textsuperscript{+6.2}} \\
\bottomrule
\end{tabular}
\end{adjustbox}
\label{tab:mainresult} 
\vspace{-4mm}
\end{table*}

\subsection{Experimental Setup}

\noindent\textbf{Baselines.}
We evaluate G2F-RAG on representative open-source LMMs, including InternVL3.5 \cite{wang2025internvl3_5}, LLaVA-Video \cite{zhang2024llavavideo}, and Qwen2.5-VL \cite{Qwen2.5-VL}, across different parameter scales. For broader context, we report results on strong open-source families such as QwenVL \cite{qwen3vl}, InternVL \cite{chen2024internvl2_5, zhu2025internvl3}, MiniCPM-V \cite{yao2024minicpm, yu2025minicpm45}, VideoLLaMA 3 \cite{zhang2025videollama3}, and VILA-1.5 \cite{lin2023vila}, as well as commercial models GPT-4o \cite{gpt4o} and Gemini 1.5 Pro \cite{gemini1.5}. We further compare against state-of-the-art video RAG baselines, including Video-RAG \cite{luo2025videorag} and Vgent \cite{shen2025vgent}. All models use public checkpoints and official inference configurations.

\noindent\textbf{Benchmarks.}
We choose 8 datasets that span complementary skills: general multi-task ability with MVBench \cite{li2024mvbench}, MMBench-Video \cite{fang2024mmbench}, and VideoMME \cite{fu2025video}; temporal and spatial reasoning with TempCompass \cite{liu2024tempcompass} and VSIBench \cite{yang2025vsibench}; open world setting required external knowledge with WildVideo \cite{yang2025wildvideo} and VideoMMMU \cite{hu2025videommmu}; and long-form reasoning with MLVU \cite{zhou2025mlvu}. We adopt the no-subtitle setting for VideoMME and the single-turn setting for WildVideo.

\noindent\textbf{Implementation Details.}
The multi-agent pipeline uses LMMs without additional training. The offline graph-construction agent employs GPT-4o \cite{gpt4o} (API Version: \texttt{gpt-4o-2024-05-13}). Difficulty routing and subgraph extraction use GPT-4o-mini \cite{gpt4omini} (API Version: \texttt{gpt-4o-mini-2024-07-18}). The reasoning frame is produced by a lightweight renderer Graphviz \cite{ellson2001graphviz}. Prompts are provided in the Appendix. All experiments run on NVIDIA A100 40GB GPUs under the official decoding and preprocessing settings of each model.

\begin{figure*}[t]
    \centering
    \includegraphics[width=\linewidth]{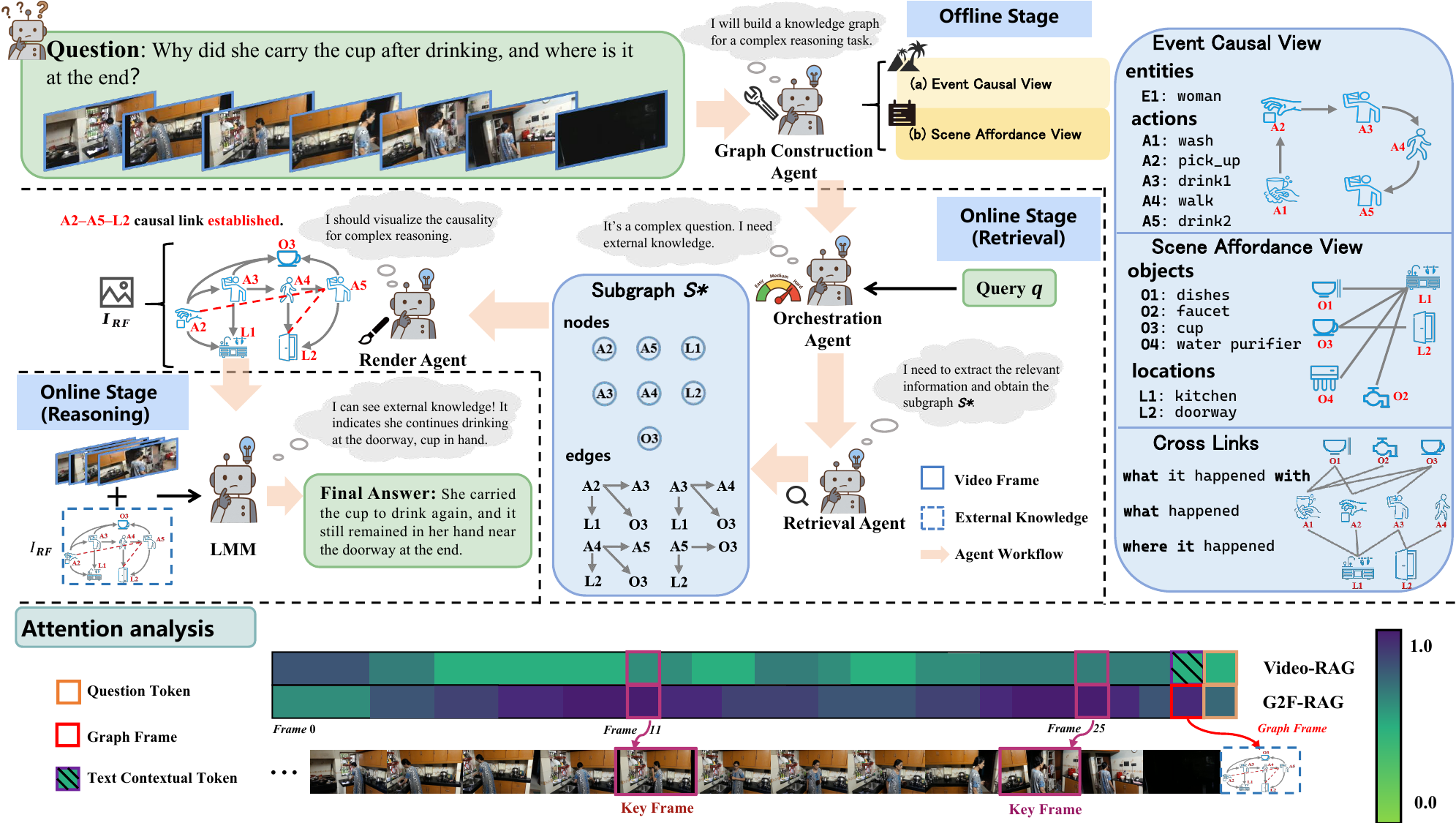}
    \vspace{-6.5mm}
    \caption{\textbf{Case study.} We builds an offline full graph, retrieves a compact subgraph online, renders one reasoning frame, and appends it to the video for reasoning. Further analysis show attention concentrates on key video frames and the graph frame.}
    \label{fig:case}
    \vspace{-6mm}
\end{figure*}

\subsection{Main Result}
As shown in Table~\ref{tab:mainresult}, G2F-RAG delivers consistent and significant gains across diverse LMM families and parameter scales without retraining the backbones. These results show that graph-to-frame fusion is portable across architectures and additive to existing capabilities.

\noindent\textbf{Analysis among Different Tasks.}
Grouping benchmarks by skill reveals a clear pattern. On temporal–spatial reasoning, G2F-RAG strengthens frame-level focus and yields large gains, exemplified by a +7.4\% improvement on VSIBench with Qwen2.5-VL-7B. On external-knowledge and open-world settings, visual-space delivery is even more beneficial, as shown by 7.1\% on WildVideo with InternVL3.5-8B. General multi-task evaluations also improve reliably: VideoMME typically rises by about +5–6\% across backbones (e.g., Qwen2.5-VL-7B +5.5\%), while MMBench-Video shows small but consistent positive shifts on its 0–3 scale. These results indicate that converting retrieved knowledge into a single reasoning frame transfers across models and task types.

\begin{table}[h]
\centering
\vspace{0.5mm}
\caption{\textbf{Performance comparison with sota RAG methods.} We compare 2 sota video RAG methods (Video-RAG, Vgent) with our G2F-RAG. G2F-RAG consistently outperforms alternatives. The best results are \textbf{bold.} All scores are average accuracy (\%).}
\vspace{-1mm}
\begin{adjustbox}{width=\linewidth,center}
\renewcommand{\arraystretch}{1.2}
\setlength{\tabcolsep}{1.5mm}
\begin{tabular}{lcccc}
\toprule
\textbf{Method} & \textbf{Size} & \textbf{MLVU} & \textbf{WildVideo} & \makecell{\textbf{VideoMME} \\ \textbf{(w/o sub)}}  \\
\midrule
LLaVA-Video & 7B & 69.5 & 53.4 & 63.7 \\
~~~~+ Video-RAG \cite{luo2025videorag} & 7B & 71.3 & 48.5 & 64.8 \\
~~~~+ Vgent \cite{shen2025vgent} & 7B & 72.5 & 51.6 & 66.7 \\
\rowcolor{blue!4} ~~~~+ G2F-RAG (Ours) & 7B & \textbf{75.5} & \textbf{57.0} & \textbf{64.5} \\
\midrule
Qwen2.5-VL & 7B & 68.8 & 51.3 & 65.1 \\
~~~~+ Video-RAG \cite{luo2025videorag} & 7B & 63.4 & 47.2 & 60.5 \\
~~~~+ Vgent \cite{shen2025vgent} & 7B & 72.1 & 50.1 & 68.9 \\
\rowcolor{blue!4} ~~~~+ G2F-RAG (Ours) & 7B & \textbf{73.4} & \textbf{55.4} & \textbf{70.6} \\
\bottomrule
\end{tabular}
\end{adjustbox}
\label{tab:rag} %
\vspace{-7mm}
\end{table}

\noindent\textbf{Analysis among Different Model Sizes.}
Smaller and medium models achieve larger relative gains, while larger models also benefit. Absolute improvements for 4B and 7B models often fall in the range of three to seven points. The gains for 8B and 14B are slightly smaller but remain steady, especially on knowledge- and structure-intensive tasks. The pattern indicates that the effect comes from representation alignment rather than capacity. Feeding knowledge as frames aligns with the video attention space and reduces the loss from cross-modal competition, which is orthogonal to size.

\noindent\textbf{Comparison with Different RAG Method.}
Text-based Video-RAG frequently degrades performance, for example Qwen2.5-VL-7B on MLVU, WildVideo, and VideoMME (w/o sub) changes from 68.8\% to 63.4\%, 51.3\% to 47.2\%, and 65.1\% to 60.5\%. Vgent mitigates overload with structured retrieval and validation but still falls short of graph-to-frame fusion. On Qwen2.5-VL-7B, MLVU and VideoMME (w/o sub) rise from 72.1\% and 68.9\% with Vgent to 73.4\% and 70.6\% with G2F-RAG. On LLaVA-Video-7B, WildVideo increases from 53.4\% to 57.0\%. These findings suggest that, what to retrieve matters, but how to represent and fuse knowledge with the video is decisive.

\noindent\textbf{Case Study.}
As shown in Figure~\ref{fig:case}, the offline graph encodes actions, objects, and locations with event–scene links. The orchestrator agent marks the query as hard; retrieval returns a minimal subgraph that ties the pick-up and repeated-drinking actions with the doorway location and the cup entity. The rendered reasoning frame depicts this causal and spatial chain and is appended as the last frame. Finally, the LMM correctly outputs the grounded answer based multimodal augmented input. The attention heatmap highlights focus on the key temporal segments and the reasoning frame under G2F-RAG, illustrating how visual-space fusion provides auditable evidence and reduces cognitive load. 

\vspace{-1mm}
\subsection{Ablation Study}
\vspace{-1mm}

\begin{table}[t]
\centering
\caption{\textbf{Ablation Study on Qwen2.5-VL-7B.} Representation/Delivery: G2J-RAG injects a textual graph json; G2F-RAG renders the retrieved subgraph into a frame and fuses it visually. Frame Placement/Count: End-1 appends a single frame at the end of the video sequence; other placements or more frames reduce performance. Frame Visual Style: Minimal uses icons and short labels; text-heavy or single-aspect styles degrade results. Routing/Fallback: Off applies RAG to all queries and hurts easy cases. Graph \& Retrieval: Coarse removes intent and affordances; Full-Loose selects larger subgraphs; Full-NoExt disables external web tools; Full-Compact selects a compact subgraph.}
\vspace{-2mm}
\begin{adjustbox}{width=\linewidth,center}
\renewcommand{\arraystretch}{1.15}
\setlength{\tabcolsep}{2mm}
\begin{tabular}{llccc}
\toprule
\textbf{Ablation} & \textbf{Variant} & \textbf{MLVU} & \textbf{WildVideo} & \makecell{\textbf{VideoMME}\\ \textbf{(w/o sub)}} \\
\midrule
\multicolumn{5}{l}{\textit{Baseline}} \\
\midrule
& Qwen2.5-VL-7B & 68.8 & 51.3 & 65.1 \\
\midrule
\multicolumn{5}{l}{\textit{Representation / Delivery}} \\
\midrule
& G2J-RAG & 66.2 & 48.8 & 63.0 \\
\rowcolor{blue!5} & G2F-RAG (Ours) & \textbf{73.4} & \textbf{55.4} & \textbf{70.6} \\
\midrule
\multicolumn{5}{l}{\textit{Frame Placement / Count}} \\
\midrule
& Start-1 & 72.4 & 54.3 & 69.4 \\
& Mid-1 & 67.9 & 49.8 & 64.0 \\
& End-4 & 69.0 & 51.6 & 66.0 \\
& End-2 & 72.4 & 54.6 & 69.8 \\
\rowcolor{blue!5} & End-1 (Ours) & \textbf{73.4} & \textbf{55.4} & \textbf{70.6} \\
\midrule
\multicolumn{5}{l}{\textit{Frame Visual Style}} \\
\midrule
& Text-Heavy & 68.4 & 50.5 & 66.2 \\
& Topology & 71.0 & 53.3 & 69.4 \\
& Causality & 70.5 & 53.0 & 69.2 \\
& Icons-Only & 69.6 & 52.4 & 68.5 \\
\rowcolor{blue!5} & Minimal (Ours) & \textbf{73.4} & \textbf{55.4} & \textbf{70.6} \\
\midrule
\multicolumn{5}{l}{\textit{Routing / Fallback}} \\
\midrule
& Off & 69.9 & 52.0 & 66.8 \\
& On (no FB) & 71.2 & 53.7 & 69.0 \\
\rowcolor{blue!5} & On + FB (Ours) & \textbf{73.4} & \textbf{55.4} & \textbf{70.6} \\
\midrule
\multicolumn{5}{l}{\textit{Graph \& Retrieval}} \\
\midrule
& Coarse & 70.2 & 52.1 & 68.8 \\
& Full-Loose & 71.6 & 53.5 & 69.7 \\
& Full-NoExt & 71.0 & 52.4 & 69.3 \\
\rowcolor{blue!5} & Full-Compact (Ours) & \textbf{73.4} & \textbf{55.4} & \textbf{70.6} \\
\bottomrule
\end{tabular}
\end{adjustbox}
\label{tab:unified_ablation}
\vspace{-6.5mm}
\end{table}

\noindent\textbf{Representation and Delivery.}
We compare textual json injection (G2J-RAG) with visual-space fusion (G2F-RAG). Both variants use the same retrieved subgraph; only the delivery differs. Rendering the subgraph as a single reasoning frame consistently outperforms textual json concatenation, notably on VideoMME (w/o sub) with an absolute gain of +7.6\% over G2J-RAG and +5.5\% over the no-RAG baseline, and on WildVideo with +6.6\% over G2J-RAG. On MLVU, G2F-RAG improves by +7.2\% relative to G2J-RAG. These results indicate that visual delivery avoids the long retrieved context that dilutes attention and make the consistent video augmented reasoning.

\noindent\textbf{Frame Placement and Count.}
We vary where and how many reasoning frames are inserted while fixing retrieval and rendering. Appending a single frame at the end (End-1) is best across datasets. Mid-sequence insertion disrupts temporal aggregation and yields the largest drop, for example MLVU 73.4\% to 67.9\% and VideoMME 70.6\% to 64.0\%. Placing the frame at the start is less harmful but still below the default, likely due to a primacy effect that biases the model before it has formed a stable video representation. Increasing the number of end-appended frames raises the visual token budget and slightly erodes accuracy, as seen with End-2 on VideoMME (69.8\%) and End-4 on MLVU (69.0\%). A single end frame provides the best balance between saliency and non-interference.

\noindent\textbf{Frame Visual Style.}
We evaluate five renderings for the same subgraph. ``Minimal" uses icons and short labels with compact layout. ``Text-Heavy" uses dense textual descriptions. ``Topology" keeps spatial relations and removes causal links. ``Causality" keeps causal chains and removes spatial topology. ``Icons-Only" removes all text. The {Minimal design (icons plus short labels) performs best, while dense text reintroduces context burden and degrades WildVideo from 55.4\% to 50.5\%. Removing structural cues also hurts: on MLVU, Topology and Causality variants reduce accuracy from 73.4\% to 71.0\% and 70.5\%. Icons-Only further confirms that a small amount of text is needed for disambiguation. The guideline is to keep diagrams compact, retain both spatial topology and directed causality, and avoid verbosity.

\noindent\textbf{Routing and Fallback.}
We evaluate when to invoke G2F-RAG. Forcing the frame for all queries (Off) hurts easy cases, yielding 66.8\% on VideoMME and 69.9\% on MLVU. Enabling routing without fallback(FB) improves results but remains vulnerable to occasional misrouting, for example 69.0\% on VideoMME. Routing with a conservative fallback (On + FB) achieves the best accuracy on all datasets, 70.6\% on VideoMME and 73.4\% on MLVU, by reserving framing for questions that benefit from external knowledge while protecting easy ones.

\noindent\textbf{Graph and Retrieval.}
We further ablate the offline graph and the online retrieval. ``Coarse" removes intent and affordances from the graph. ``Full-Loose" keeps the full graph but selects larger subgraphs. ``Full-NoExt" keeps the full graph but disables external tools during construction. ``Full-Compact" is our default with the full graph and compact subgraph selection. Removing intent and affordances lowers MLVU from 73.4\% to 70.2\% and VideoMME from 70.6\% to 68.8\%, showing that these fields capture useful preconditions and functionality. Using larger subgraphs modestly reduces precision, for example on VideoMME 69.7\% compared with 70.6\%. Disabling external tools during construction weakens world knowledge and reduces WildVideo from 55.4\% to 52.4\%. Our default setting pairs a rich offline graph with compact, relevance-focused retrieval and visual delivery.

To further assess robustness of G2F-RAG, we deliberately feed incorrect and adversarial reasoning frames and observe that, because the prompt consistently prioritizes grounding in the original video while our method contributes only a single frame, the overall performance drop is almost negligible, more details are in the Appendix.

\vspace{-1mm}
\section{Conclusion}
\vspace{-1mm}
We addressed a central bottleneck in retrieval-augmented video reasoning: how external knowledge is represented and fused with the backbone. We introduced Graph-to-Frame RAG (G2F-RAG), a training-free paradigm that converts retrieved structured knowledge into a single reasoning frame and delivers it in the visual space. To the best of our knowledge, this is the first work to deliver retrieved knowledge in the visual space for video reasoning. Instantiated with a two-stage multi-agent workflow, G2F-RAG reduces the cross-modal competition and token burden, and provides an auditable evidence trail. Extensive experiments across datasets and backbones show consistent gains and indicate that visual-space fusion is a more compatible path for augmenting video LMMs than unbounded textual context.

\noindent \textbf{Acknowledgement.} This work is supported by the Young Scientists Fund of the Hunan Natural Science Foundation (Grant No.2024JJ6474), the Youth Independent Innovation Science Fund Project of NUDT (Grant No.ZK24-08).

{
    \small
    \bibliographystyle{ieeenat_fullname}
    \bibliography{main}
}


\end{document}